\documentclass[runningheads]{llncs}
\pdfoutput=1
\usepackage{booktabs, multirow}
\usepackage{amsmath,amssymb} 
\usepackage{color}
\DeclareMathOperator*{\argmax}{arg\,max}

\usepackage[dvipsnames]{xcolor}
\usepackage{bm}
\usepackage{cleveref}
\usepackage{tikz}
\usepackage{rotating}
\usepackage{wrapfig}
\usepackage{pbox}
\usepackage{adjustbox}

\crefname{section}{sec.}{sec.}
\Crefname{section}{Sec.}{Sec.}
\crefname{figure}{fig.}{fig.}
\Crefname{figure}{Fig.}{Fig.}
\Crefname{equation}{eq.}{eq.}
\Crefname{equation}{Eq.}{Eq.}

\begin{document}
\title{Reinforced Coloring for End-to-End \\ Instance Segmentation} 
\titlerunning{Reinforced Coloring for End-to-End Instance Segmentation}

%
\author{Tuan Tran Anh\inst{1} \and
Khoa Nguyen-Tuan\inst{1} \and
Tran Minh Quan\inst{1} \and
Won-Ki Jeong\inst{2}}

%

\authorrunning{Tuan et al.}

%

\institute{Ulsan National Institute of Science and Technology (UNIST) \\
\email{\{anhtuanhsgs,ntkhoa,quantm\}@unist.ac.kr} \and 
Korea University \\
\email{wkjeong@korea.ac.kr}}

\maketitle              

\begin{abstract}
Instance segmentation is one of the actively studied research topics in computer vision in which many objects of interest should be separated individually. While many feed-forward networks produce high-quality segmentation on different types of images, their results often suffer from topological errors (merging or splitting) for segmentation of many objects, requiring post-processing. Existing iterative methods, on the other hand, extract a single object at a time using discriminative knowledge-based properties (shapes, boundaries, etc.) without relying on post-processing, but they do not scale well. To exploit the advantages of conventional single-object-per-step segmentation methods without impairing the scalability, we propose a novel iterative deep reinforcement learning agent that learns how to differentiate multiple objects in parallel. Our reward function for the trainable agent is designed to favor grouping pixels belonging to the same object using a graph coloring algorithm. We demonstrate that the proposed method can efficiently perform instance segmentation of many objects without heavy post-processing. 
\end{abstract}

\section{Introduction}\label{s:intro}
\begin{figure}[t]
    \centering
    \includegraphics[width=1\linewidth]{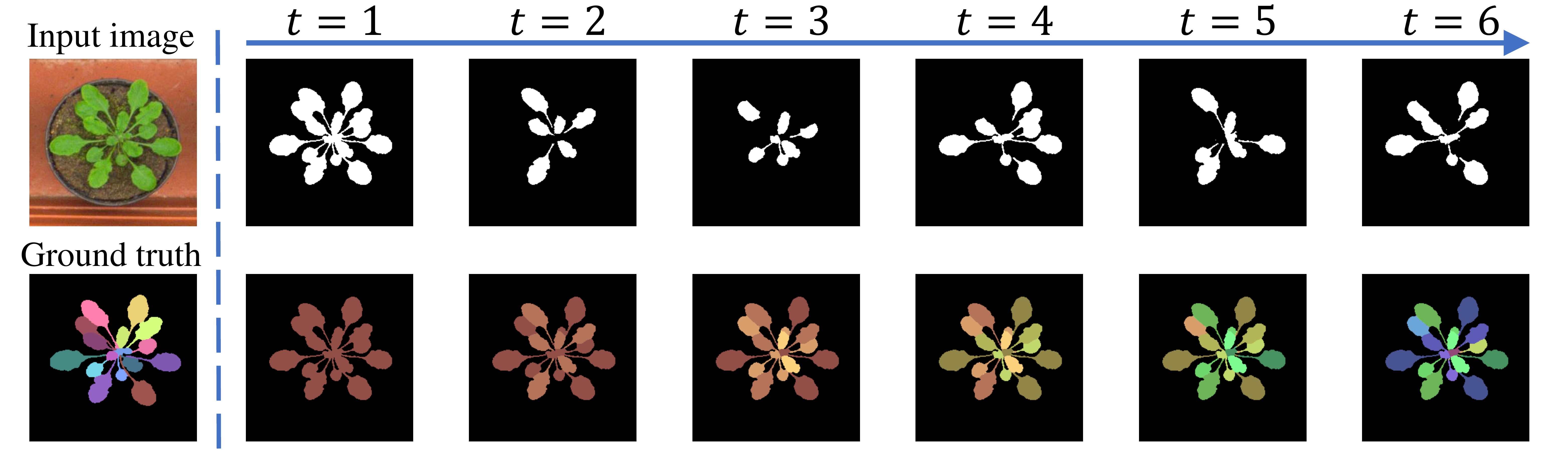}
    \caption{Illustration of our agent's coloring process. Our agent takes the input image and sequentially segments multiple objects at a time, as shown on the right side of the dashed line. The upper row shows the agent's action maps, which are also the binary maps of multiple instances. The row below shows the color maps which, sequentially change with each new action map. %
    The action map at each step is a binary digit map representing the segmentation label. %
    For example, we have the color labels of \{0 and 1\} at step 1, the color labels of \{00, 01, 10 and 11\} at step 2, etc \label{Fig:intro}}
\end{figure}
Recent advances in deep reinforcement learning (DRL) has archived human-level performance on complicated tasks that previously required human control and decision making~\cite{mnih2015human,lillicrap2015continuous,silver2016mastering}.
%
Given that the training reinforcement learning agent learns tasks in a human-like way (from experience via trial and error), the early success of DRL mainly focused on mimicking human tasks, such as playing games.
More recently, there have also been successful attempts to apply DRL in conventional computer vision tasks, such as image processing~\cite{furuta2019fully,li2018a2}. 

Instance segmentation is a challenging computer vision problem that assigns instance labels to pixels to separate objects, which is crucial for understanding a complex scene. 
Many existing instance segmentation methods arebased on complicated graphical models with deep neural networks (e.g., convolutional neural network [CNN] or recurrent neural network [RNN]))~\cite{zhang2015monocular,zhang2016instance,ren17recattend}. 
However, instance segmentation also involves decision tasks (i.e., how to assign labels to pixels), which is more complicated than conventional (semantic) object segmentation. 
Recent work by Araslanov \textit{et al.} aimed to addressed this issue by employing reinforcement learning for the sequential object detection and segmentation task~\cite{araslanov2019actor}. 
%
%

%
%
While sequential object segmentation methods like those of Araslanov~\textit{et al.} and Ren~\textit{et al.}~\cite{ren17recattend} have shown promising results on image with a small number of objects, 
their sequential methods, which segments one object at a time, are not efficient when the number of objects is large.
To address this problem, we propose a novel end-to-end instance segmentation method using reinforcement learning. 
Unlike the method where a single agent handles an object as seen in Araslanov~\textit{et al.}, our coloring agent consists of multiple pixel-level agents (as in Furuta~\textit{et al.}~\cite{furuta2019fully}) working concurrently to 
differentiate multiple objects in a sequential, end-to-end fashion (fig.~\ref{Fig:intro}).
%
To enable multiple instances to be labeled concurrently, we formulate and solve an iterative graph binary coloring problem. 
Using the asynchronous advantage actor-critic (A3C) algorithm, our agents are trained to choose the $t$-th bit value in binary representation of the label at the step $t$ of the coloring process. 
%
%
Pixel-label agents try to take actions (0 or 1) that are either matching or different at one point throughout the coloring process, depending on whether the instances are same or different. 

To the best of our knowledge, this is the first end-to-end instance segmentation that uses reinforcement learning.
We demonstrate the performance and scalability of the proposed method on several open source datasets, such as 
KITTI~\cite {geiger2012we}, CREMI~\cite{cremi_dataset}, and CVPPP~\cite {minervini2016finely}
and compare our results with the other iterative methods.
%
We demonstrate that our method can efficiently handle images with many objects of various shapes while still maintaining a competitive segmentation quality.

\section{Related Work}\label{s:related}

%
In this section, we briefly overview the recent advances in image segmentation methods, which are closely related to the instance segmentation problem.%

\textbf{Knowledge-based segmentation approaches:} Conventionally, prior \linebreak knowledge can be used to incorporate to a representation (e.g., a computational graph where pixels become nodes and the quantitative relationship between them form edges). 
Solving the min-cut and max-flow in this relationship can partition an image into discriminative regions (or segmentation)~\cite{boykov2006graph}. 
%
%
The key idea in these approaches is to construct a proper distance metric between pixels so that they can be grouped into segments where the total number of partitions can be either deterministic or not~\cite{achanta2010slic}. 
However, hand-crafted prior knowledge from those clusters is not always aligned to the goal of segmentation and left a room for improvement. 
%
%

\textbf{Supervised learning approaches:}
The invention of Fully Convolutional Neural Network (FCN)~\cite{long2015fully} and its variations, such as U-net~\cite{ronneberger2015u} with different backbones~\cite{he2016deep,huang2017densely} and different types of skip-connection~\cite{quan2016fusionnet,jegou2017one}, have achieved a big success in segmentation tasks. 
Moreover, one can focus on the loss function design that makes it possible for a cluster to collapse by itself into one region and push other clusters far away~\cite{de2017semantic}. %
%
Another direction for solving the instance segmentation task is to produce segmentation in a sequential prediction manner. 
Ren~\textit{et al.}~\cite{ren17recattend} utilize a recurrent neural network to perform step-by-step performing attention then segmentation the mask of a single object. 
This approach returns a good segmentation map for the image and also accurately returns the number of object, but does not scale well for many objects. 
%
%
The advantage of supervised learning approaches 
is that the level of hierarchical order of segmentation can be obtained directly from the data without complicated hand-crafted rules, but most methods are still sequential. 
%
%
%
%
%

\textbf{Reinforcement learning approaches:}
Since Mnih \textit{et al.}~\cite{mnih2015human} introduced their seminal work, an increasing number of complex tasks that are challenged by machine intelligence due to its complex sequences of decision making processes have been solved by reinforcement learning~\cite{silver2016mastering,kempka2016vizdoom,lillicrap2015continuous}. 
It is natural for one to seek to make use of the recent advancements in reinforcement learning and apply them to solve the problem in the computer vision domain.
For example, Furuta~\textit{et al.} presented an efficient way to train an asynchronous actor critic agent (A3C), which is called PixelRL~\cite{furuta2019fully}, that uses the decision making per pixel for the denoising problem. 
%
To investigate how those sequential steps can form the segmentation solution pipeline, people have constructed the segmentation procedure as a Markov Decision Process (MDP) and attempted to solve it by leveraging several state-of-the-art algorithms in reinforcement learning. 
Araslanov~\textit{et al.}~\cite{araslanov2019actor} formulated the instance-aware segmentation problem into a sequential object detection-segmentation action decision making process. 
Gwangmo~\textit{et al.}~\cite {song2018seednet} made an agent that uses the random walk segmentation algorithm ~\cite{grady2006random} with human interaction input to sequentially extract the region of interest. However, it is still lacking a method that can segment multiple objects at a time in a sequential manner.  

\section{Graph Coloring Approach}
\label{GraphColoring}
\begin{figure}[t]
    \centering
    \includegraphics[width=1\linewidth]{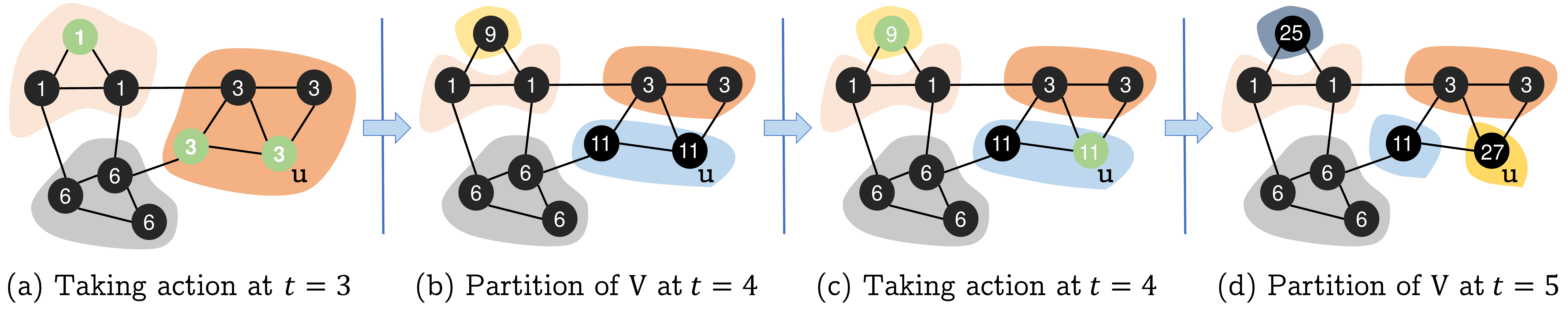}
    \caption{
    An example of the binary coloring process by the agent. 
    The number on each vertex represents its color label, and the green (or black) vertex during the action selection phase represents 
    the action value of 1 (or 0). %
    For example, at step $t=4$, the vertex \textbf{u} with the color label of 11 ($C^{(4)}(V)[\textbf{u}]=11$, see (b)), it chooses the action value of 1 ($\mathcal{F}^{(4)}(V, C^{(4)}(V))[\textbf{u}]=1$ (green vertex), see (c)). %
    Then, the color label of $\textbf{u}$ at $t=5$ becomes $11+2^{4}\cdot1=27$ (see (d)) \label{Fig:BinaryColoring}}
\end{figure}

\subsection{Problem Formulation}
\label{sec:ProbFormulate}

In this work, we formulate the instance segmentation problem into a multi-step graph coloring problem, similar to D. Gómez \textit{et al.}~\cite{gomez2007graph}. 
Given that image \textit{I} consists of the set of pixels $V=\{v_1, v_2, ..., v_n\}$. A segmentation of \textit{I} partitions $V$ into $P=\{P_1, P_2, ..., P_m\}$, where each $v_i$ belongs to exactly one $P_j$ for $1 \leq i \leq n$ and $1 \leq j \leq m$. 
By constructing the set of edges $E$ and graph $G=(V,E)$ from $V$, we can formulate the instance segmentation problem into a graph coloring problem. 
For each image \textit {I}, we want to find a color (label) mapping that assigns a color to every pixel, $C: V \to \{0, 1, 2, ..., c-1\}$, %
that satisfies the following constraints. Given a graph $G=(V,E)$ and a ground truth partitioning $\hat{P}$ of $V$: $C(u) = C(v)$ if $\exists P_j \in \hat{P}$ s.t $u \in P_j$ and $v \in P_j$ and $(u,v) \in E$; $C(u) \neq C(v)$ if $\nexists P_j \in \hat{P}$ s.t $u \in P_j$ and $v \in P_j$ and $(u,v) \in E$. 
%
%
Then, the image segmentation problem is finding a proper function $\mathcal{F}$ that maps a set of graphs $\mathcal{G}$ to the set of color mapping that satisfies the above constrains.


Since the task of finding an optimal $\mathcal{F}$ is an NP-Hard problem~\cite{cormen2009introduction}, 
so 
we find the approximation of $\mathcal{F}$ using an iterative binary coloring process.
%
We begin by letting $C^{(t)}(V)$ be the color mapping of $G$ at time step $t$; and defining coloring action $a^{(t)}=\mathcal{F}^{(t)}(V, C_{t}(V))$, where $\mathcal{F}^{(t)}$ maps $V$ to $\{0, 1\}^N$, and  $N$ is the size of $V$. Each $v$ of $V$ is mapped to $0$ or $1$ though $\mathcal{F}^{(t)}$. 
$\mathcal{F}^{(t)}(V, C_{t}(V))[v]$ denotes the mapped value of $v$, and the color of $v$ at time step $t+1$ is computed as follows (we illustrate this function in Figure~\ref{Fig:BinaryColoring}):
\begin{equation}
    C^{(t+1)}(V)[v]=C^{(t)}(V)[v]+2^{t}\mathcal{F}^{(t)}(V, C^{(t)}(V))[v]   
\end{equation}
Here $C^{(t+1)}(V)[v]$ returns the color mapping of a single vertex $v$ in $V$.
If T is the maximum number of coloring steps, then we have a $T$-step approximation function of $\mathcal{F}$, which maps $V$ to $\{0, 1, 2, ..., 2^{T-1}\}$. 
It can be seen that $\mathcal{F}^{(t)}(V, C^{(t)}(V))[v]$ is assigned to the $t$-th digit in the binary representation of color of $v$. 
%
%
%
%

\subsection{A3C and PixelRL} 
For the coloring problem, %
we can naturally think of a multi-agent system where each agent is in charge of taking action $a^{(t)}=\mathcal{F}^{(t)}$ that changes $C^{(t)}$ for a single vertex of $V$. 
Asynchronous actor critic (A3C) is one of the policy gradient algorithms 
that has demonstrated high performance for discrete action space decision-making problems~\cite {mnih2016asynchronous}. 
In this work, we employed the method introduced by Furuta~\textit{et al.}~\cite{furuta2019fully} which uses an efficient technique for a multi-agent system (PixelRL) which works well 
with A3C. 
%
%

An image \textit{I} has a set of pixels $V=\{v_{1}, v_{2}, ..., v_{N}\}$ in PixelRL problem setting. 
Each $v_{i}$ has a corresponding state $s_{i}^{(t)}$ at time step $t$. 
A pixel-level agent with policy $\pi(a_{i}^{(t)} \mid s_{i}^{(t)})$ is assigned to each pixel $v_{i}$. 
State $s^{(t+1)}=(s_{1}^{(t+1)}, s_{2}^{(t+1)}, ..., s_{N}^{(t+1)})$ and reward $r^{(t)}=(r^{(t)}_{1}, r^{(t)}_{2}, ..., r^{(t)}_{N})$ are obtained from the environment by taking action $a^{(t)}=(a^{(t)}_{1}, a^{(t)}_{2}, ..., a^{(t)}_{N})$, $a^{(t)}_{i} \in \mathcal{A}$. 
In our work, $\mathcal{A}$ has only two values, $0$ and $1$, which represent the binary digit value of label color.
The agents try to 
maximize the mean of their total expected reward: 
\begin{equation}
    \pi^{*} = \argmax_{\pi}  E_{\pi} (\sum_{t=0}^{\infty}\gamma^{t}\bar{r}^{(t)}), \\
\end{equation}
\begin{equation}
    \bar{r}^{(t)}={1\over{N}}\sum_{i=1}^{N}r_{i}^{(t)}
\end{equation}
where $\bar{r}^{(t)}$ is the mean rewards $r_{i}^{t}$. 
At each time step $t$, with state $s^{(t)}$, PixelRL agent computes the value function $\mathcal{V}(s^{(t)})$ and policy function ${\pi}(s^{(t)})$.
$\mathcal{V}(s^{(t)})$ estimates the expected reward an agent can get from the state $s^{(t)}$, which implies how good the state $s^{(t)}$ is. 
Loss functions $L_{value}^{i}$ of $\mathcal{V}$ and $L_{policy}^{i}$ of ${\pi}$ for a single agent at pixel $v_i$ are computed as follows: 
	\begin{equation} \label{eq:ValueLoss}
	    L_{value}^{i}= (R^{(t)}_{i} - \mathcal{V}(s^{(t)}_{i}))^2
	\end{equation}
	\begin{equation} \label{eq:PolicyLoss}
	    L_{policy}^{i} = -log (\pi(a^{(t)}_{i} \mid s^{(t)}) A(s^{(t)}_{i}))
	\end{equation}
where $A^{(t)} = R^{(t)}_{i} - \mathcal{V}(s^{(t)}_{i})$ is the advantage function, which shows how good the action $a^{(t)}_{i}$ at step $t$ is compared to the expected return. 
At each time step $t$, gradients for value loss and policy loss are computed and used to update the parameters of $\mathcal{V}$ and $\pi$.
In PixelRL, a convolutional neural network is used to compute $\mathcal{V}$ and $\pi$; $\mathcal{V}$ and $\pi$ have the same dimensions as the input image $s^{(t)}$. 
For more information, see Furuta \textit{et al.}~\cite{furuta2019fully}.

\subsection{Coloring Agent}
\begin{figure}[t]
    \centering
    \includegraphics[width=1\linewidth]{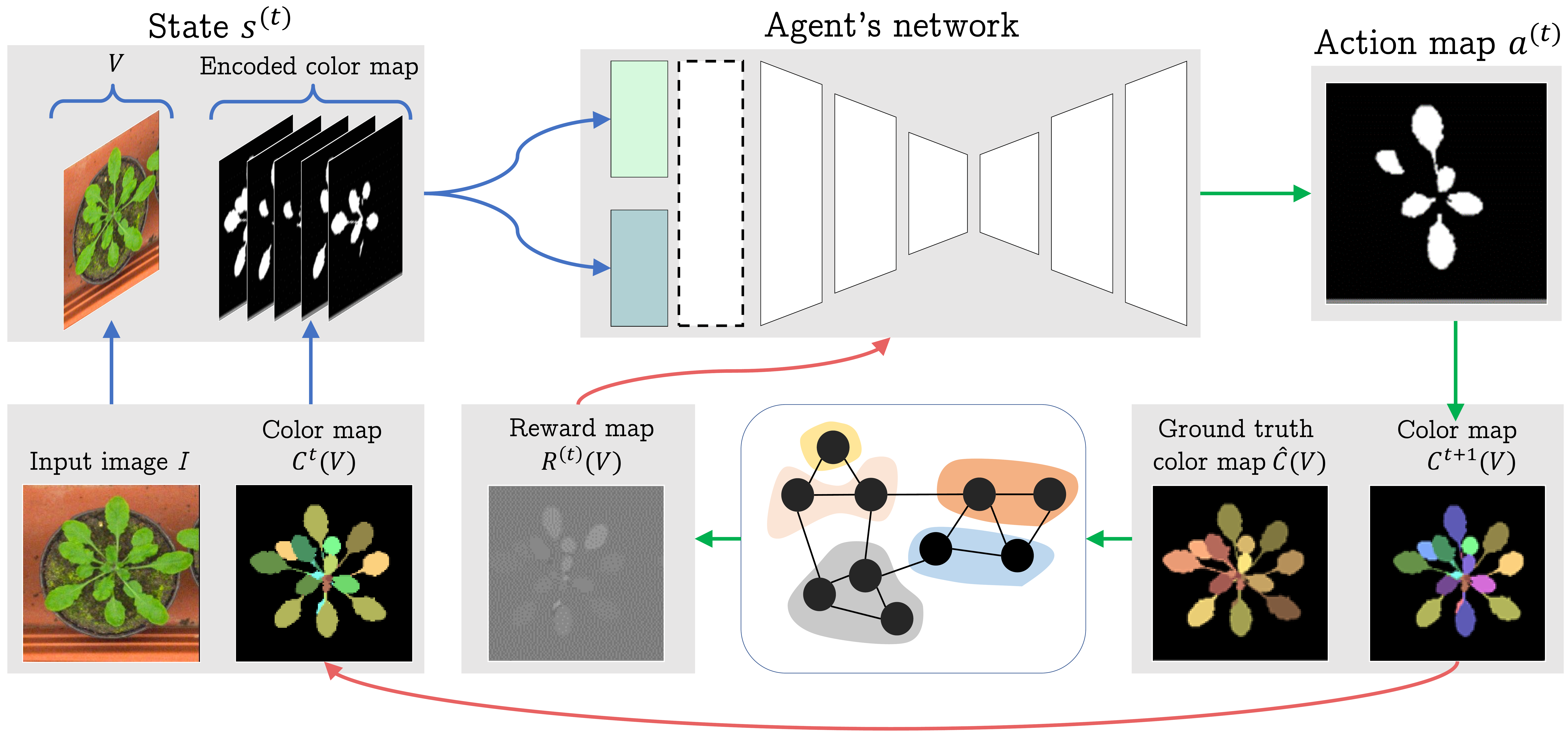}
    \caption{Overview of our coloring agent. %
    The state of the agent comprises the sets of pixels $V$ (input image \textit{I}) and the binary representation of label $C^{t}(V)$. %
    Blue arrows indicate input paths, which lead the current state to the agent network. %
    The input image \textit{I} and the binary color map 
    go into different modules %
    (the two colored boxes inside Agent's network) to be concatenated and processed by a CNN. %
    Green arrows indicate the action-related paths. %
    After getting a new color map $C^{(t+1)}(V)$ by updating the action map $a^{(t)}$, a graph algorithm will take the ground truth label $\hat{C}(V)$ and $C^{(t+1)}(V)$ to produce reward map $R^{(t)}(v)$. %
    Red arrows indicate update-related paths where the network and state are updated using a new reward and a color map, respectively
    \label{Fig:Architecture}}
\end{figure}
Our coloring agent processes the state $s^{(t)}$ at time step $t$ to produce a binary map of $N$ actions, and each action makes a change for a single pixel label.
%
The action map is also a binary mask of multiple-object segmentation.
We formulate the Markov Decision Process for the instance segmentation problem with the tuple of state, action, and reward. 
Figure~\ref{Fig:Architecture} shows an overview of the agent architecture.
This section will explain these three terms in detail. \\\\ 
\noindent
\textbf{State:}
Function $\mathcal{F}^{(t)}$ takes the input, which is a set of vertices $V$ (the image \textit{I}), and its color map $C^{(t)}(V)$. 
Given image \textit{I} of size $H \times W \times K$, the representation of input $V$ and $C^{(t)}$ for $\mathcal{F}$ here are the image \textit{I} and its binary encoded $T$-channels color map. 
Thus, the state of an agent is an image of size $H \times W \times (K + T)$, where T is the number of coloring steps. 
Here, $K$ is the channels of image \textit{I} and $T$ is the number of binary digits of color map. \\\\
\textbf{Actions:}
Action map $a^{(t)}$ that resulted from $\mathcal{F}^{(t)}(V, C^{(t})[V]$ is a binary image of size $H \times W$ as defined in Section~\ref{sec:ProbFormulate}. The action map $a^{(t)}$ at the time step $t$ can be seen as a segmentation map of several objects at that time step. \\\\
%
\noindent
\textbf{Rewards:}
For each pixel, to get the reward map $r^{(t)}$, we need to construct the set of edges $E$ between pixels from the ground truth label $\hat{P}$ and $C_{t}$.
The goal of the reward function for each action is to give reasonable feedback for the actions that cause pixels to have different colors (splitting actions) and the actions that keep pixels having the same colors (merging actions). 
We divide the reward function into three major components, one that encourages the splitting actions, another that encourages merging actions, and the third one that classifies between foreground and background labels. 
Figure~\ref{Fig:EdgeBuilding} illustrates the edges construction phase for the computation of reward function. 
To make the reward function more instance-focused, the edges are constructed only between foreground pixels while the separation between foreground and background is done specially at first step with a designated reward component. 
We denote $\hat{C}(v_i)$ be the ground truth label of pixel $v_i$, and $\hat{C}(v_i)=0$ when $v_i$ of the background only.
$\hat{P}(v_i)$ is the ground truth segment that contains $v_i$ (ie. $\hat{P}(v_i)=\{v_j \mid \hat{C}(v_i)=\hat{C}(v_j)\}$). \\\\
\textbf{Reward for predicting background-foreground:}\\
We design a reward function just to segmenting between background region and foreground region. 
By doing this, the background pixels do not need to compare with each other (especially when the image has complex background structures like in electron microscope (EM) images). 
The reward function for separating foreground and background is defined as follows:
\begin{equation} \label{eq:reward_locator} 
    \!\begin{aligned}
        R_{BF}^{(t)}(v) = \begin{cases} r_{bg} &\mbox{if } C^{(t+1)}(u)=0, \hat{C}(v)=0 \\ 
                0 &\mbox{if } \hat{C}^{(t+1)}(v) \neq 0, t > 0\\
                -r_{bg} & \mbox{if } C^{(t+1)}(u) \neq 0, \hat{C}(v)=0 \\
                r_{fg} & \mbox{if } C^{(t+1)}(u)=1, \hat{C}(v) \neq 0, t=0 \\
                -r_{fg} & \mbox{if } C^{(t+1)}(u)=0, \hat{C}(v) \neq 0, t=0 \end{cases} 
    \end{aligned}	
\end{equation}
Here, we set $r_{bg}$ and $r_{fg}$ to be 
the percentage of the foreground and background areas to the entire area, respectively.
%
In the first step, we made the problem to be only differentiating between foreground and background. 
After that, our agent separates objects while maintaining the background prediction. 
Thus, the foreground components ($r_{fg}$ and $-r_{fg}$) are given only in the first step (at $t=0$). 
\\\\
\textbf {Reward for spliting actions:}
%
%
%
By constructing edges between pixels of different ground truth (GT) segments, we wish to compare their color and give feedback to the actions that return the color mapping. 
Given the positive integer $r$, the edge list constructed using $r$ is denoted as $E^{r}$. 
%
A directed edge originating from $u$ to $v$ is defined as a tuple ($u,v$).
Then $(u,v)\in E^{r} $ %
if $\hat{C}(v) \neq \hat{C}(u)$ and $\hat{C}(v) \neq 0$ and $\hat{C}(u) \neq 0$ and $\exists u^{'} \in \hat{P}(u)$ s.t $d(u^{'}, v) < r$ where $d(x, y)$ is the Manhattan distance between $x$ and $y$. 
%
$r$ can be considered as the radius of segments, so we call $r$ a \textit{splitting radius}. 
Figure~\ref{Fig:EdgeBuilding}a illustrates how edges originating from $v$ are constructed using a given splitting radius $r$. 
%
%
We then define the set 
$FM_{r}^{(t)}(v)$ and $TS_{r}^{(t)}(v)$ for a pixel $v$ at time step $t$ as follows:
%
%
\begin{equation}
    \begin{aligned}
        TS_{r}^{(t)}(v)= \{u\mid (v, u)\in E^{r}, C^{(t)}(u) \neq C^{(t)}(v),\hat{C}^{(t)}(u)\neq\hat{C}^{(t)}(v)\}
    \end{aligned}
\end{equation}
\begin{equation}
    \begin{aligned}
        FM_{r}^{(t)}(v)= \{u\mid (v, u)\in E^{r}, C^{(t)}(u) = C^{(t)}(v), \hat{C}^{(t)}(u)\neq\hat{C}^{(t)}(v)\}
    \end{aligned}
\end{equation}
%
%
%
%
As outlined above, $TS_{r}^{(t)}(v)$ and $FM_{r}^{(t)}(v)$ can be represented as the set of neighborhoods of $v$ that are correctly split and incorrectly merged. 
For a pixel $v$ with radius $r$, at time step $t$  $(1 \leq t < T)$, the splitting reward $R_{S}^{(t)} (v \mid E^{r})$ is computed as follows:\\
\begin{equation}
    R_{TS}^{(t)}(v\mid E^{r}) = \frac{|TS_{r}^{(t+1)}(v)| - |TS_{r}^{(t)}(v)|} {|\{(v,u) \mid (v, u)\in E^{r}\}|}
\end{equation}
\begin{equation}
    R_{FM}^{(t)}(v\mid E^{r}) = \frac{1}{T}\frac{|FM^{(t+1)}_{r}(v)|} {|\{(v,u) \mid (v, u)\in E^{r}\}|}
\end{equation}
\begin{equation}
   R_{S}^{(t)}(v\mid E^{r}) = R_{TS}^{(t)}(v\mid E^{r}) - R_{FM}^{(t)}(v\mid E^{r})
\end{equation}
{
\setlength{\tabcolsep}{1em} 
\begin{figure}[t]
    \centering
    \begin{tabular}{cc}
        \includegraphics[width=0.3\linewidth]{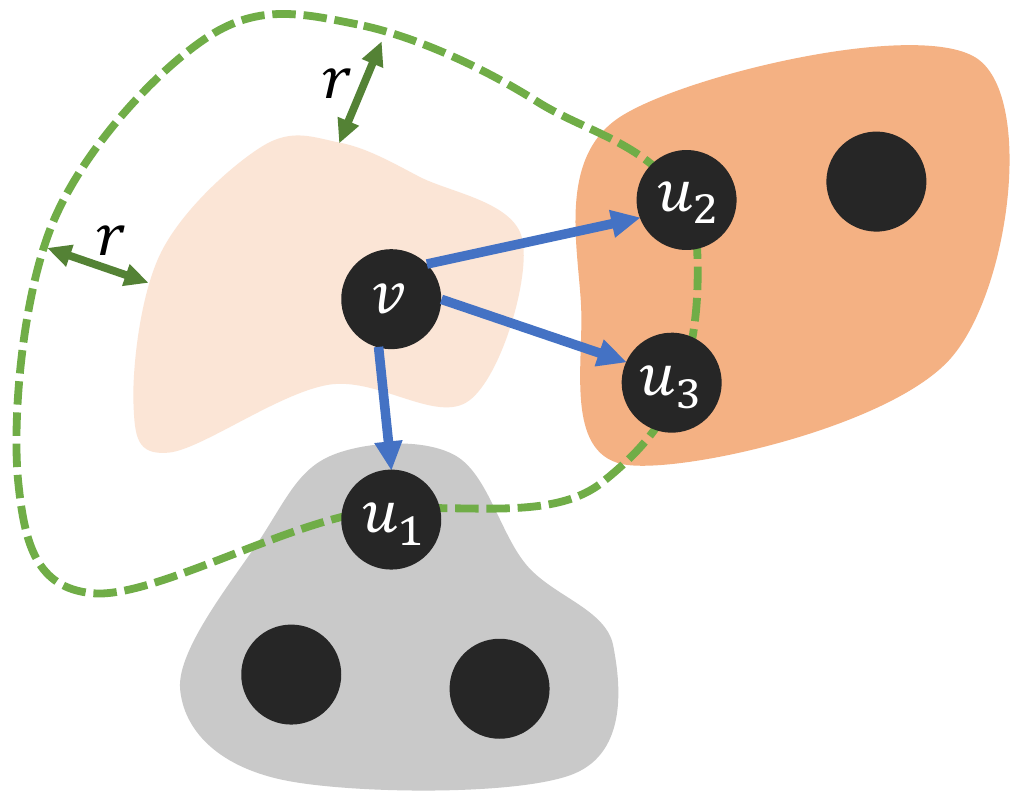} &
        \includegraphics[width=0.3\linewidth]{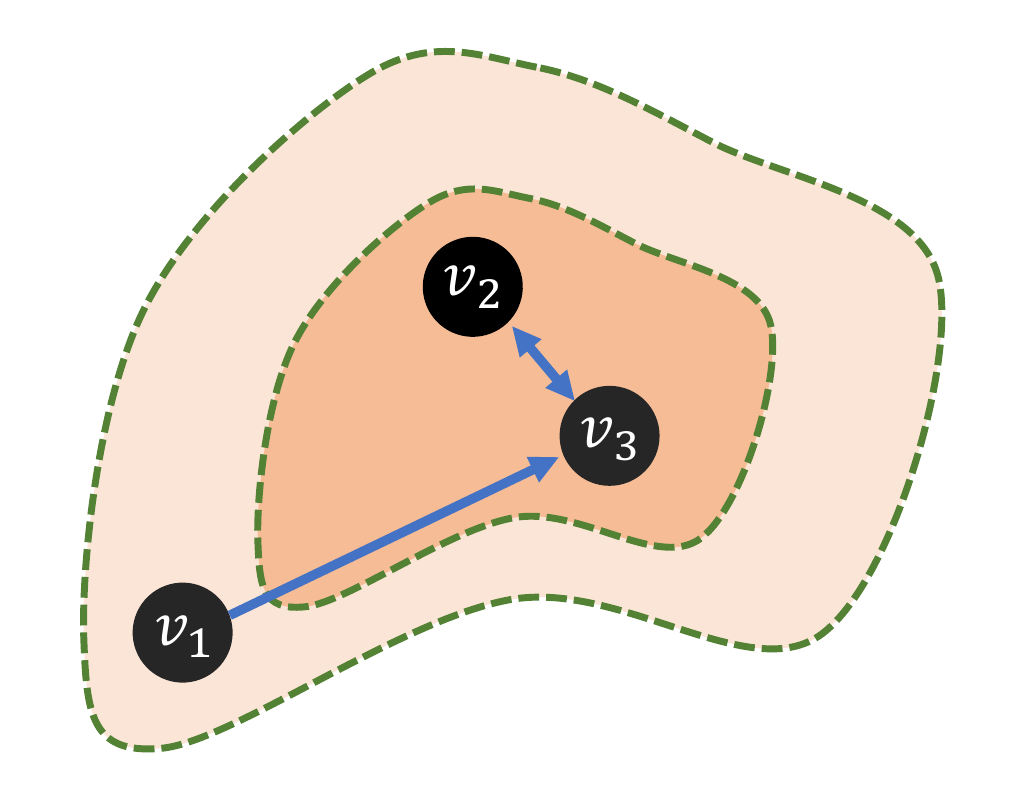} \\
        (a) Edge construction for splitting & (b) Edge construction for merging \\
    \end{tabular}
    \caption{Illustration of the edge construction. 
    (a): For the splitting-related rewards, we only consider the edges connected to the ground-truth segment containing $v$ within the distance $r$. 
    %
    (b): For the merging-related rewards, pixels near the boundary of a segment only compare the color of itself with pixels of the inner region while every pair of pixels in the inner region are connected bi-directional edges
    \label{Fig:EdgeBuilding}}
    
\end{figure}
}

%
\textbf{Reward for merging actions:} %
We construct edges between pixels in the same ground truth segment for the merging reward function. 
The reward function guides the pixel-level agents from the same ground truth segment to take the same actions.
%
%
%
%
For an object, it is more important for the pixels of the inner region to have the same color with each other than for a pixels of the outer region to have the same color with other pixels inside the object. We give a higher priority for matching color between pixels in the inner region.
Given a \textit{shrinking factor} $\alpha$ ($0 \leq \alpha \leq 1$), the inner region of a ground truth segment $\hat{P}(u)$ containing $u$ is generated by shrinking $\hat{P}(v)$ to $\hat{P}_\alpha(v)$ such that $|\hat{P}_{\alpha}(v)| < minsize$ or ${|\hat{P}_{\alpha}(v)|}/{|\hat{P}(v)|} < \alpha$. 
%
The directed edge list $E_{\alpha}$ is constructed as follows (illustration of the graph construction is in Figure. ~\ref{Fig:EdgeBuilding}):
$(u, v) \in E_{\alpha}$ if $\hat{C}(u)=\hat{C}(v)$ and $v \in \hat{P}_{\alpha} (u)$. 
We then define the set $TM_{\alpha}^{(t)}(v)$ and $FS_{\alpha}^{(t)}(v)$ %
for a pixel $v$ at time step $t$ as follow:
\begin{equation}
    \begin{aligned}
        FS_{\alpha}^{(t)}(v)= \{u\mid (v, u)\in E_{\alpha}, C^{(t)}(u) \neq C^{(t)}(v), \hat{C}^{(t)}(u)=\hat{C}^{(t)}(v)\}
    \end{aligned}
\end{equation}
\begin{equation}
    \begin{aligned}
        TM_{\alpha}^{(t)}(v)= \{u\mid (v, u)\in E_{\alpha}, C^{(t)}(u) = C^{(t)}(v), \hat{C}^{(t)}(u)=\hat{C}^{(t)}(v)\}
    \end{aligned}
\end{equation}
$FS_{\alpha}^{(t)}(v)$ and $TM_{\alpha}^{(t)}(v)$ are the set of neighborhoods of $v$ that are wrongly split from $v$ and correctly merged with $v$. 
For a pixel $v$ with shrinking factor $\alpha$ at time step $t$ ($1 \leq t \leq T$), the merging reward $R_{FS}^{(t)}(v \mid E_{\alpha})$ is computed as follows:
\begin{equation}
    R_{TM}^{(t)}(v \mid E_{\alpha}) = \frac{1}{T} \frac{|TM^{(t)}_{\alpha}(v)|} {|\{(v, u)\mid (v, u)\in E_{\alpha}\}|}
\end{equation}
\begin{equation}
    R_{FS}^{(t)}(v \mid E_{\alpha}) = \frac{|FS^{(t-1)}_{\alpha}(v)|-|FS^{(t)}_{\alpha}(v)|} {|\{(v, u)\mid (v, u)\in E_{\alpha}\}|}
\end{equation}
\begin{equation}
   R_{M}^{(t)} (v \mid E_{\alpha}) = R_{TM}^{(t)}(v \mid E_{\alpha}) - R_{FS}^{(t)}(v \mid E_{\alpha})
\end{equation}
\\\\
\textbf{Reward for pixel $v$ at time step $t$:}\\
Our reward function $R^{(t)}(v)$ for a vertex is described as follow:\\
When $1 \leq t < T $:
\begin{equation}    
    \begin{aligned}
        R^{(t)}(v) = R_{BF}^{(t)} (v) + w_{m}\sum_{E_{\alpha} \in \mathcal{G}_m} R_{M}^{(t)} (v \mid E_{\alpha}) + w_{s} \sum_{E^{r} \in \mathcal{G}_s}  R_{S}^{(t)}(v\mid E^{r})
    \end{aligned}
\end{equation}
 and when $t=0$:
 \begin{equation}    
    \begin{aligned}
        R^{(t)}(v) = R_{BF}^{(t)} (v)
    \end{aligned}
\end{equation}
 where $w_{m}$ and $w_{s}$ are weights for merging and splitting, respectively. 
 $\mathcal{G}_s$ and $\mathcal{G}_m$ are the sets of $E_{\alpha}$(s) and $E^{r}$(s) for different values of $\alpha$ and $r$, respectively. The higher the value for $w_m$ compared to $w_s$, the higher the chance that actions that keep the merged area intact will be chosen, and vice versa.

\section{Experiments and Results}
In this work, we used Attention U-Net architecture (AttU)~\cite{oktay2018attention} for the core network of our agent. %
Due to the difference between input image space and label color space, we let input image \textit{I} and the binary color map go though two different paths before merging them by concatenation as input for AttU, as shown in the overview structure (Fig.~\ref{Fig:Architecture}). %
For pre-processing modules, we use astrous spatial pooling layers. 
%
We set the discount factor with the default value of $\gamma=1.0$ and shrinking factor $\alpha=0.8$ in all the experiments.

\begin{figure}[t]
    \centering
    \includegraphics[width=1\linewidth]{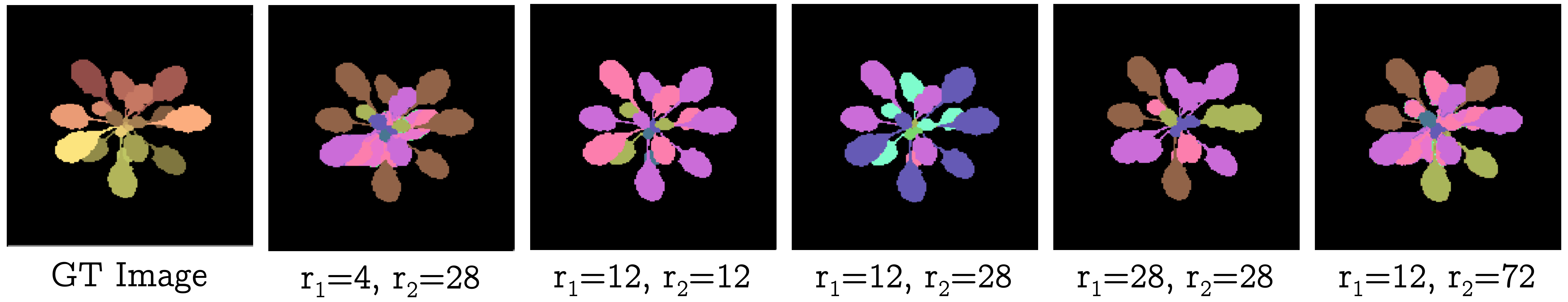}
    \caption{Different radius settings from inferencing and training on a single image 
    \label{Fig:Abation_radius}}
    
\end{figure}

\begin{figure}[t]
    \centering
    \begin{tabular}{cc}
        \includegraphics[height=3.6cm]{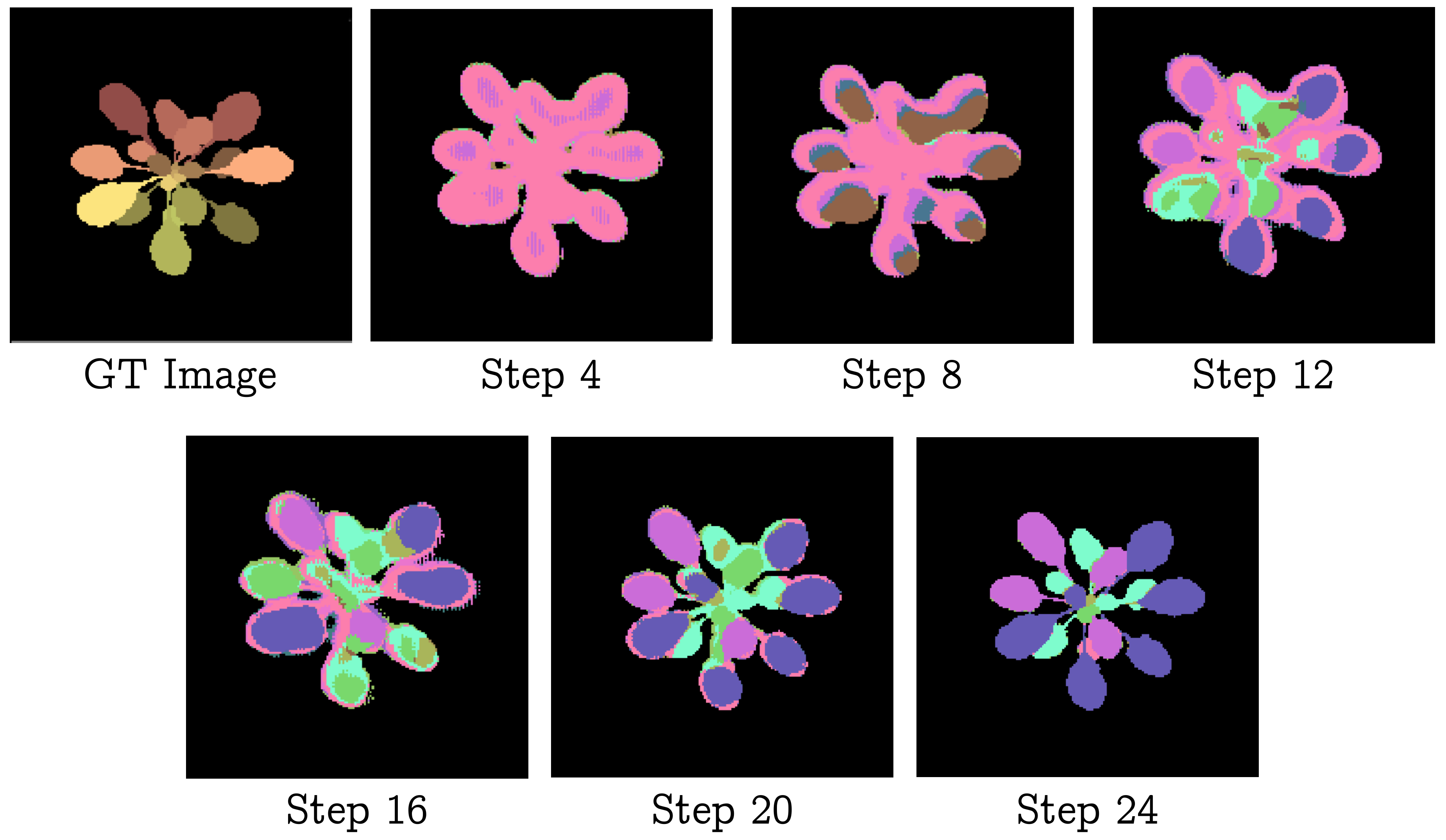}&
        \includegraphics[height=4cm]{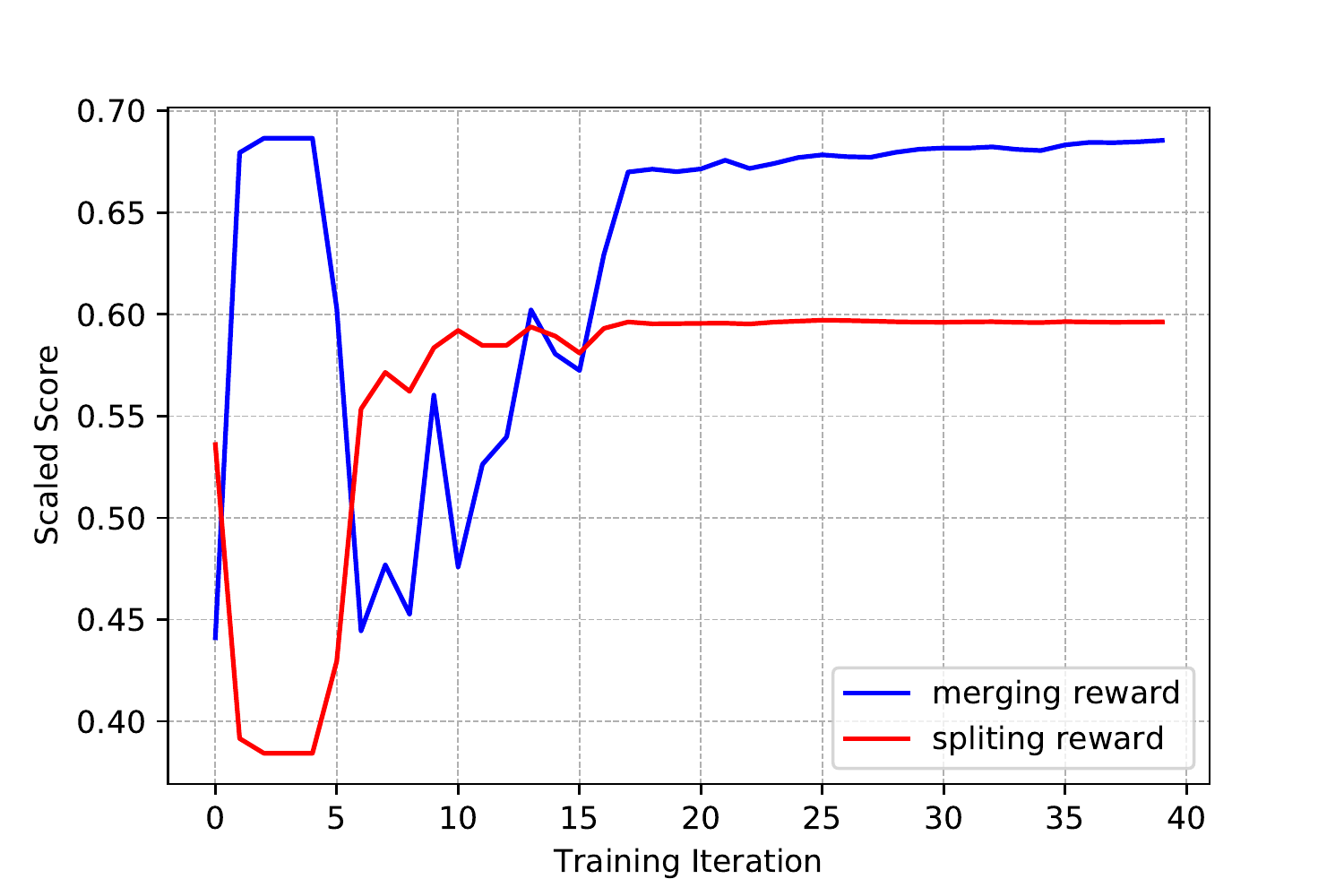} \\
        (a) & (b) \\
    \end{tabular}
    \caption{Result from single image fitting. The line graph shows scaled values of merging and splitting rewards during the training iterations. The images on the left show the GT image and the intermediate results at different training iterations 
    \label{Fig:Abation_Stage}}
\end{figure}

\begin{figure}[t]
    \centering
    \includegraphics[height=3.6cm]{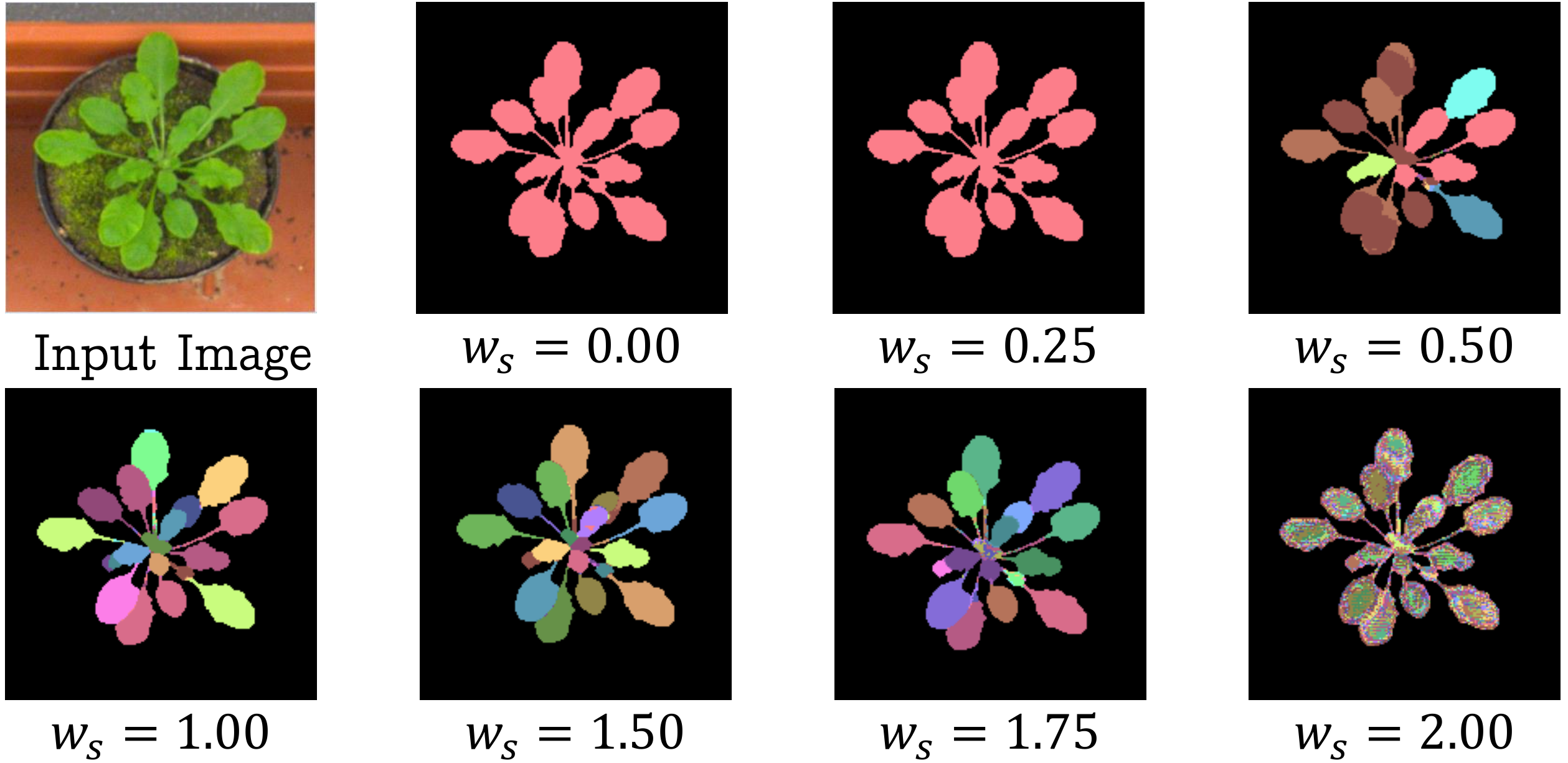}
    \caption{%
    Inference results of models trained with different splitting and merging weights setting on a test image (for all images, $w_{m}+w_{s}=2$) %
    \label{Fig:Abation_weight}}
\end{figure}

\subsection{Ablation Study}
\label{SubS:ablation}

\textbf{Splitting radius setting:} 
%
Separating objects within close proximity is more important and challenging.
%
By exploiting different levels of splitting radius $r$(s), the agent can learn to do segmentation better. 
Here, we analyze the behavior of our agent with two levels of splitting radii $r_{1}$ and $r_{2}$. The environment setting becomes simpler as we let the agent to only learn to segment a single training image (no augmentation and $w_{s}=w_{m}=1$ also).
We observed that $r_{1}=12$ and $r_{2}=28$ gave the best result among the trials (Fig.~\ref{Fig:Abation_radius}). 
While a small radii setting gives the agent enough information to differentiate close and small objects, there is no feedback for the agent to separate large and far apart instances ($r_{1}=12,r_{2}=12$). 
A large radii setting, on the other hand, gives long-distance information but also makes the task harder as the pixels have to process more ($r_{1}=28,r_{2}=28$). 
Too small ($r_{1}=4,r_{2}=28$) or too big radius ($r_{1}=12,r_{2}=72$) components can also guide the agent poorly as too small radii often contribute almost no useful information and too large radii make the task much harder. 
\\\\
\textbf{Weights for splitting and merging rewards:} %
%
We analyzed how the reward functions affect the agent by testing different sets of weights for splitting and merging rewards. 
We used 103 training images and 25 validation images of CVPPP, 
and fixed the sum of $w_{s}$ and $w_{m}$ to a constant of 2 in this experiment. 
The results using different weight settings are shown in Table~\ref{table:ablation} and Figure~\ref{Fig:Abation_weight}. %
We see that the low merge-split weight ratio does affect the segmentation quality of our reinforced coloring agent (RC) as much high merge-split weight ratio. %

{
\setlength{\tabcolsep}{1em} 
\begin{table}[t]
    \centering
    \caption{Results of the CVPPP validation set with different weights setting and comparison of two graph construction algorithms for the reward function}
    \label{table:ablation}
    \begin{tabular}{c | c c | c c | c c}
        \toprule
        Model & $w_{s}$ & $w_{m}$ & $r_1$ & $r_2$ & SBD$\uparrow$ & $\mid$DiC$\mid$$\downarrow$ \\
        \midrule
               & 0.00 & 2.00 &    &     &  21.2 &    15.6\\
               & 0.25 & 1.75 &    &     &  21.2 &    9.44\\
               & 0.50 & 1.50 &    &     &  70.7 &    2.72\\ 
        RC     & 1.00 & 1.00 & 12 &  28 &  85.2 &    1.40\\
               & 1.50 & 0.50 &    &     &  87.3 &    1.34\\
               & 1.7  & 0.25 &    &     &  81.4 &    1.44\\
               & 2.00 & 0.00 &    &     &  5.60 &    92.8\\
        \bottomrule
    \end{tabular}
\end{table}
}

We observed that during the training and exploration for better decision making, our agent reaches the easier stage first (maximization of merging reward) then gradually finds actions that differentiate objects (maximization of splitting reward) (Fig.~\ref{Fig:Abation_Stage}a). 
During the latter stages, maximizing splitting rewards may come with the cost of merging reward at some point (Fig.~\ref{Fig:Abation_Stage}b). 
Thus, it is necessary that ${{w_s}\over{w_m}} > 1$ for the trade-off of splitting and merging rewards. 
%
Based on this result, 
for all the experiments discussed in the following sections,
we choose $w_{m}=1.0$ and $w_{s}=1.5$ for a little higher incentive to the agent for exploring splitting actions. 
While $\alpha$ is always set to $0.8$ to relax the learning difficulty of instance border areas, we use different splitting radius $r$ for different datasets. 

\subsection {CVPPP Dataset}
{
\setlength{\tabcolsep}{10pt}
\begin{table}[t]
    \centering
    \caption{\emph {Segmentation quality of CVPPP testset}. The metrics we used are Symmetric Best Dice (SBD) and absolute Difference in Counting ($\mid$DiC$\mid$)}
    \label{table:results_cvppp_test}
    \begin{tabular}{lcc}
        \toprule
        Model & SBD$\uparrow$ & $\mid$DiC$\mid$$\downarrow$ \\
        \midrule
        RIS~\cite{romera2016recurrent} & 66.6 & 1.1\\ 
        MSU~\cite{scharr2016leaf} & 66.7 & 2.3\\
        Nottingham~\cite{scharr2016leaf} & 68.3 & 3.8\\
        IPK~\cite{pape20143} & 74.4 & 2.6\\
        DLoss~\cite{de2017semantic} & 84.2 & 1.0\\ 
        E2E~\cite{ren17recattend} & 84.9 & 0.8\\
        AC-Dice~\cite{araslanov2019actor} & 79.1 & 1.12\\
        \midrule
        Ours $\textit{(RC)}$   & 80.0  & 1.36 \\
        \bottomrule
    \end{tabular}
\end{table}
}

The Computer Vision Problems in Plants Phenotyping (CVPPP) dataset is one of the popular datasets used for assessing the performance of instance segmentation algorithms. 
%
%
We used the A1 dataset, which consists of 128 training images and 33 testing images. 
%
%
We resized the images down to $176\times176$ pixels (the original size was $530\times500$ pixels)  
and used two levels of splitting radius $r_{1}=12$ and $r_{2}=28$ as discussed in Section~\ref{SubS:ablation}.
\begin{figure}[t]
    \centering
    \includegraphics[width=1\linewidth]{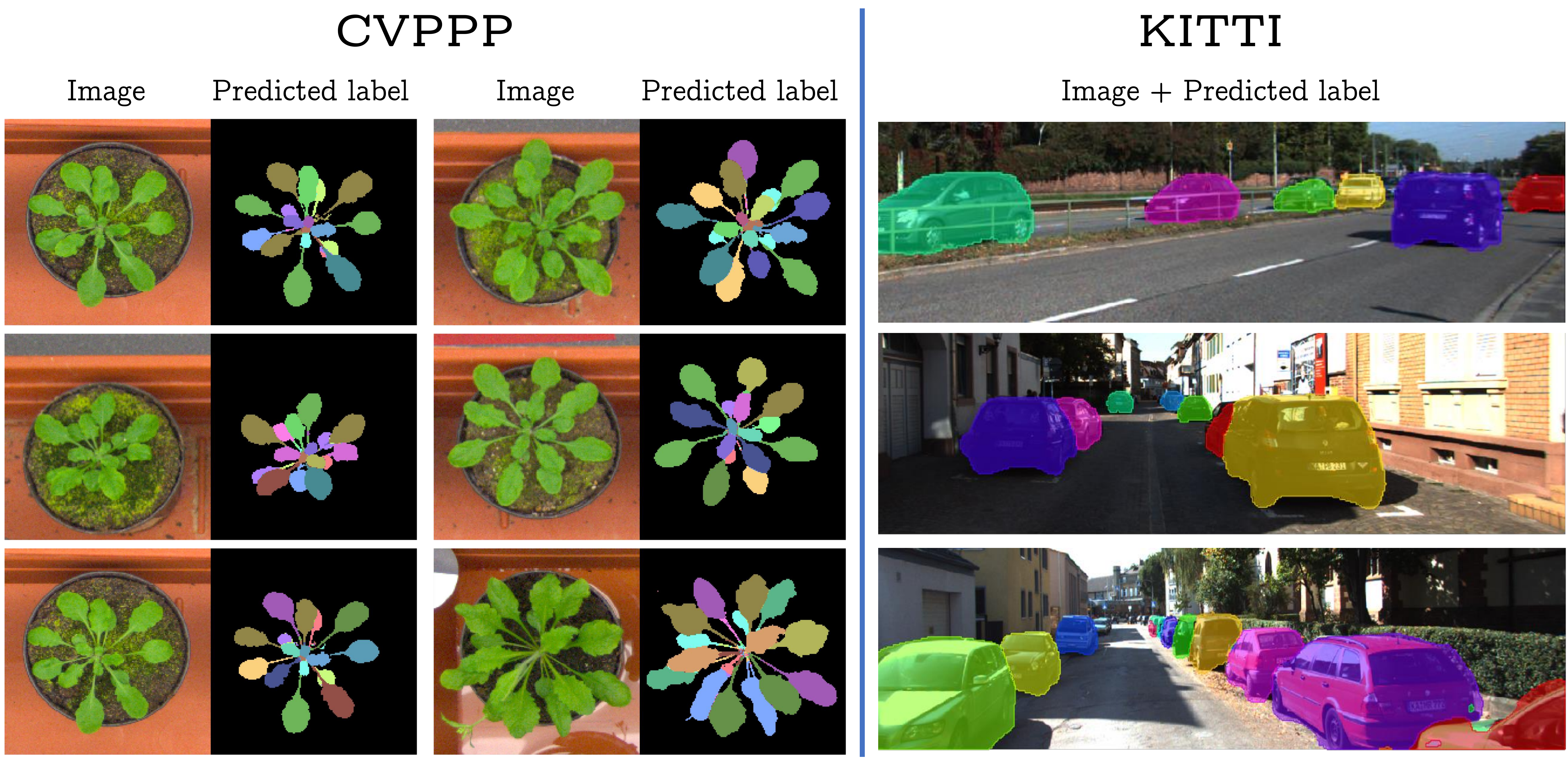}
    \caption{Results of CVPPP and KITTI test dataset. In this figure, we relabeled the label of KITTI results while keeping the result of CVPPP as it is 
    \label{fig:Result}}
\end{figure}
%
We allow our agent to use the same label color for objects that are far apart from each other.
For the sake of the evaluation, for all the data sets, before the evaluation of segmentation accuracy, the predicted label map is further post-processed with resizing (upscaling to the original size), removing small segments, and re-indexing labels. The evaluation is done on the original size of the data.
%
%
The quality of the segmentation is measured in the Symmetric Best Dice (SBD) and the absolute Difference in Counting ($|$DiC$|$) measurements. 
The checkpoint used for the evaluation is selected from the one that has the best $|$DiC$|$ score.
Comparing our results with Ren~\textit{et al.} (E2E) and Araslanov~\textit{et al.} (AC-Dice), while our $|$DiC$|$ score is slightly lag behind, our agent produces segmentation quality on par with their methods (see Table~\ref {table:results_cvppp_test}). Figure~\ref{fig:Result} shows that our agent can segment the leaves also handle occlusions well.
\subsection{KITTI Dataset}
{
\setlength{\tabcolsep}{5pt}
\begin{table}[t]
    \centering
    \caption{\emph{Segmentation quality on KITTI testset.} We evaluate our method (\textit{RC}) in terms of mean weighted (MWCov) and unweighted (MUWCov) coverage, average false positive (AvgFP), and false negative (AvgFN) rates.}
    \label{table:results_kitti_test}
    \begin{tabular}{lcccc}
        \toprule
        Model & MWCov$\uparrow$ & MUCov$\uparrow$ & AvgFP$\downarrow$ & AvgFN$\downarrow$ \\
        \midrule
        DepthOrder~\cite{zhang2015monocular} & 70.9 & 52.2 & 0.597 & 0.736\\
        DenseCRF~\cite{zhang2016instance} & 74.1 & 55.2 & 0.417 & 0.833\\
        AngleFCN+D~\cite{uhrig2016pixel} & 79.7 & 75.8 & 0.201 & 0.159\\
        E2E~\cite{ren17recattend} & 80.0 & 66.9 & 0.764 & 0.201\\
        AC-BL-Trunc~\cite{araslanov2019actor} & 72.2 & 50.7 & 0.393 & 0.432\\
        AC-IoU~\cite{araslanov2019actor} & 75.6 & 57.3  & 0.338 & 0.309\\
        \midrule
        Ours $\textit{(RC)}$  & 77.0 & 68.5  & 0.249 & 0.128\\
        \bottomrule
    \end{tabular}
\end{table}
}
We also assess the performance of our method on the KITTI car segmentation dataset. 
We use the same 3712 images for training, 144 images for validation and 120 images for testing as in~\cite{araslanov2019actor,ren17recattend}. 
In KITTI dataset, the training labels generated from~\cite{papandreou2015weakly} are in a coarse resolution but the testing and validation images are in a high resolution, which makes the problem challenging~\cite{papandreou2015weakly,chen2016monocular}. 
%
We downsampled the training images to $160\times480$ (originally $256\times1024$ pixels). 
Since vehicles in KITTIS are often distributed sparsely in the images and their number is also small, we set our agent to do 4-step coloring. 
In this data, we use two levels of the radius ($r_{1}=8$ and $r_{2}=32$). 
The post-processing setting for evaluation is the same as the setting we used with CVPPP. 

The metrics used for evaluation of this data are the mean weighted coverage (MWCow), the mean unweighted coverage loss (MWCow), the average false positive rate (AvgFP), and the average false negative rate (AvgFN). 
MUCow measures the instance-wise IoU for each GT instance averaged over the image, while MWCow is the average of IoUs of predicted labels matched with GT instances weighted by the size of GT instances~\cite{ren17recattend}. 
AvgFP is the fraction of predicted label segments that do not have matched GT segments. 
AvgFN is the fraction of GT label segments that do not have a matched label prediction. 
Our result is shown in Table~\ref{table:results_kitti_test}, which illustrates that our AvgAP and AvgFN scores are better than Ren \textit{et al.} and Araslanov \textit{et al.}'s single-object-per-step approaches. 
Previous comparison with result from Figure~\ref{fig:Result} demonstrate that our method can learn and generalize well from the incomplete annotation.

\subsection {CREMI Dataset}
\begin{figure}[t]
    \centering
    \includegraphics[height=5cm]{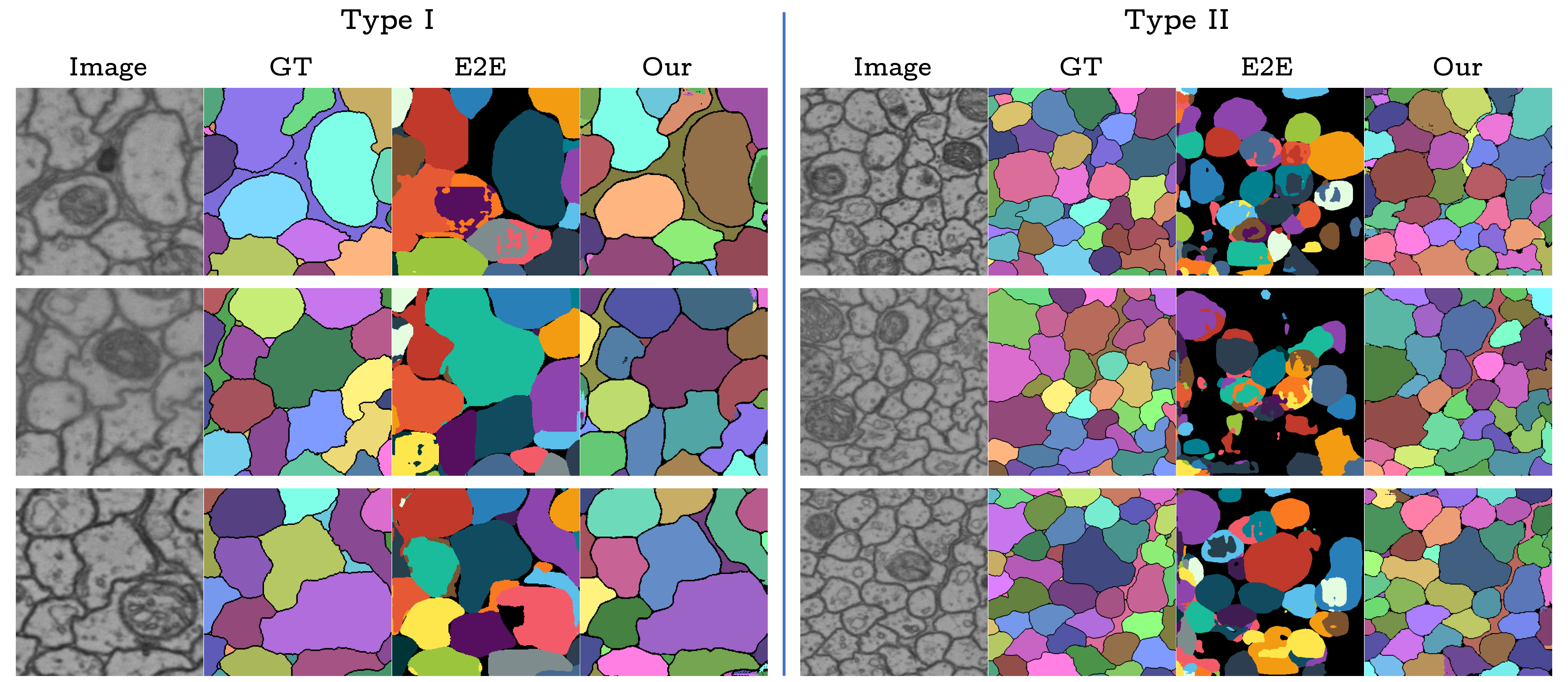}
    \caption{Results on CREMI test datasets with different input sizes
    \label{fig:E2E_cremi}}
\end{figure}

{
\setlength{\tabcolsep}{4.5pt}
\begin{table}[t]
    \centering
    \caption{\emph{Segmentation quality of CREMI testset.} We compare our method with Ren~\textit{et al}'s method on CREMI dataset in terms of VOI-split, VOI-merge, adapted rand index (ARand) and average inference time per image (avg. time)}
    \label{table:results_E2E_cremi_test}
    \begin{tabular}{l|c|cccc}
        \toprule
        Model & Data type & avg. time(ms) & VOI-split$\downarrow$ & VOI-merge$\downarrow$ & ARand$\downarrow$ \\
        \midrule
        E2E~\cite{ren17recattend} & \multirow{2}{*}{Type I} & 514.83 & 0.772 & 0.544 & 0.276 \\
        Ours & & 171.76 & 0.412 & 0.113 & 0.07\\
        \midrule
        E2E~\cite{ren17recattend} & \multirow{2}{*}{Type II} & 910.46 & 1.178 & 3.082 & 0.660 \\
        Ours & & 186.97 & 0.379 & 0.230 & 0.095 \\
        
        \bottomrule
    \end{tabular}
\end{table}
}

CREMI is an electron microscope image dataset in which many cell objects are densely packed. We chose this dataset to demonstrate both the segmentation quality and the scalability of our method.
We used  a padded version of CREMI dataset A, which has 125 sections of images of $1250\times1250$ pixels. 
%
We prepared two versions of the dataset from the original one: type I and type II. Dataset type I has patches of size $256\times256$ pixels and each patch has 24 cells on average (maximum is 40). Dataset type II has patches of size $448\times448$, and each patch has on average 65 cells (80 at most). For each type, we randomly extract 103 patches from the first 100 sections for the training set and 25 patches from the last 25 sections for the test set. 
Training images were downsampled to $224\times224$.
%

%
Quality metrics used in this experiment are a Variation of Information (VOI-split, VOI-merge), adapted RAND error (ARAND), and mean inference time per patches (.avg time). %
Figure~\ref{fig:E2E_cremi} and Table~\ref{table:results_E2E_cremi_test} show that our agent can capture better shape and size of cells. 
While E2E can find and segment densely packed cells (although not perfect) in type I images, the method easily loses its tracking of cells (large regions are classified as background) in type II images. 
CREMI images contain many cells of complex structures and varying sizes as well as noise and occlusions, 
which makes the problem more challenging for the attention-then-segmentation approach like E2E. 
Our method, on the other hand, can effectively handle densely packed many objects by separating multiple objects in parallel via iterative binary segmentation (i.e., graph coloring).
%
The average inference time (Avg.time) is also measured (post-processing time is included). While the inference time of E2E linearly increases with the number of objects, our average inference time stayed constant, which shows the superior scalability of our method.
%
%
%

%



\section{Conclusion}
In this paper, we introduced a novel per-pixel label assignment method for end-to-end 
instance segmentation based on a graph coloring approach. 
We proposed a reward function that gives meaningful feedback for each pixel to decide its label index iteratively. 
%
%
Based on the evaluation of three datasets (KITTI, CVPPP, and CREMI), we demonstrated that the proposed method is effective for instance segmentation of many objects. 
In the future, we plan to conduct rigorous performance the evaluation on large-scale multiple-object segmentation.

\bibliographystyle{splncs04}
\bibliography{refs}

\begin{thebibliography}{10}
\providecommand{\url}[1]{\texttt{#1}}
\providecommand{\urlprefix}{URL }
\providecommand{\doi}[1]{https://doi.org/#1}

\bibitem{achanta2010slic}
Achanta, R., Shaji, A., Smith, K., Lucchi, A., Fua, P., S{\"u}sstrunk, S.: Slic
  superpixels. Tech. rep. (2010)

\bibitem{araslanov2019actor}
Araslanov, N., Rothkopf, C.A., Roth, S.: Actor-critic instance segmentation.
  In: Proceedings of the IEEE Conference on Computer Vision and Pattern
  Recognition. pp. 8237--8246 (2019)

\bibitem{boykov2006graph}
Boykov, Y., Funka-Lea, G.: Graph cuts and efficient nd image segmentation.
  International journal of computer vision  \textbf{70}(2),  109--131 (2006)

\bibitem{chen2016monocular}
Chen, X., Kundu, K., Zhang, Z., Ma, H., Fidler, S., Urtasun, R.: Monocular 3d
  object detection for autonomous driving. In: Proceedings of the IEEE
  Conference on Computer Vision and Pattern Recognition. pp. 2147--2156 (2016)

\bibitem{cormen2009introduction}
Cormen, T.H., Leiserson, C.E., Rivest, R.L., Stein, C.: Introduction to
  algorithms (2009)

\bibitem{cremi_dataset}
CREMI: Miccai challenge on circuit reconstruction from electron microscopy
  images (2016), \url{https://cremi.org/}

\bibitem{de2017semantic}
De~Brabandere, B., Neven, D., Van~Gool, L.: Semantic instance segmentation with
  a discriminative loss function. arXiv preprint arXiv:1708.02551  (2017)

\bibitem{furuta2019fully}
Furuta, R., Inoue, N., Yamasaki, T.: Fully convolutional network with
  multi-step reinforcement learning for image processing. In: Proceedings of
  the AAAI Conference on Artificial Intelligence. vol.~33, pp. 3598--3605
  (2019)

\bibitem{geiger2012we}
Geiger, A., Lenz, P., Urtasun, R.: Are we ready for autonomous driving? the
  kitti vision benchmark suite. In: 2012 IEEE Conference on Computer Vision and
  Pattern Recognition. pp. 3354--3361. IEEE (2012)

\bibitem{gomez2007graph}
G{\'o}mez, D., Montero, J., Y{\'a}{\~n}ez, J., Poidomani, C.: A graph coloring
  approach for image segmentation. Omega  \textbf{35}(2),  173--183 (2007)

\bibitem{grady2006random}
Grady, L.: Random walks for image segmentation. IEEE Transactions on Pattern
  Analysis \& Machine Intelligence (11),  1768--1783 (2006)

\bibitem{he2016deep}
He, K., Zhang, X., Ren, S., Sun, J.: Deep residual learning for image
  recognition. In: Proceedings of the IEEE conference on computer vision and
  pattern recognition. pp. 770--778 (2016)

\bibitem{huang2017densely}
Huang, G., Liu, Z., Van Der~Maaten, L., Weinberger, K.Q.: Densely connected
  convolutional networks. In: Proceedings of the IEEE conference on computer
  vision and pattern recognition. pp. 4700--4708 (2017)

\bibitem{jegou2017one}
J{\'e}gou, S., Drozdzal, M., Vazquez, D., Romero, A., Bengio, Y.: The one
  hundred layers tiramisu: Fully convolutional densenets for semantic
  segmentation. In: Proceedings of the IEEE Conference on Computer Vision and
  Pattern Recognition Workshops. pp. 11--19 (2017)

\bibitem{kempka2016vizdoom}
Kempka, M., Wydmuch, M., Runc, G., Toczek, J., Ja{\'s}kowski, W.: Vizdoom: A
  doom-based ai research platform for visual reinforcement learning. In: 2016
  IEEE Conference on Computational Intelligence and Games (CIG). pp.~1--8. IEEE
  (2016)

\bibitem{li2018a2}
Li, D., Wu, H., Zhang, J., Huang, K.: A2-rl: Aesthetics aware reinforcement
  learning for image cropping. In: Proceedings of the IEEE Conference on
  Computer Vision and Pattern Recognition. pp. 8193--8201 (2018)

\bibitem{lillicrap2015continuous}
Lillicrap, T.P., Hunt, J.J., Pritzel, A., Heess, N., Erez, T., Tassa, Y.,
  Silver, D., Wierstra, D.: Continuous control with deep reinforcement
  learning. arXiv preprint arXiv:1509.02971  (2015)

\bibitem{long2015fully}
Long, J., Shelhamer, E., Darrell, T.: Fully convolutional networks for semantic
  segmentation. In: Proceedings of the IEEE conference on computer vision and
  pattern recognition. pp. 3431--3440 (2015)

\bibitem{minervini2016finely}
Minervini, M., Fischbach, A., Scharr, H., Tsaftaris, S.A.: Finely-grained
  annotated datasets for image-based plant phenotyping. Pattern recognition
  letters  \textbf{81},  80--89 (2016)

\bibitem{mnih2016asynchronous}
Mnih, V., Badia, A.P., Mirza, M., Graves, A., Lillicrap, T., Harley, T.,
  Silver, D., Kavukcuoglu, K.: Asynchronous methods for deep reinforcement
  learning. In: International conference on machine learning. pp. 1928--1937
  (2016)

\bibitem{mnih2015human}
Mnih, V., Kavukcuoglu, K., Silver, D., Rusu, A.A., Veness, J., Bellemare, M.G.,
  Graves, A., Riedmiller, M., Fidjeland, A.K., Ostrovski, G., et~al.:
  Human-level control through deep reinforcement learning. Nature
  \textbf{518}(7540),  529--533 (2015)

\bibitem{oktay2018attention}
Oktay, O., Schlemper, J., Folgoc, L.L., Lee, M., Heinrich, M., Misawa, K.,
  Mori, K., McDonagh, S., Hammerla, N.Y., Kainz, B., et~al.: Attention u-net:
  Learning where to look for the pancreas. arXiv preprint arXiv:1804.03999
  (2018)

\bibitem{papandreou2015weakly}
Papandreou, G., Chen, L.C., Murphy, K.P., Yuille, A.L.: Weakly-and
  semi-supervised learning of a deep convolutional network for semantic image
  segmentation. In: Proceedings of the IEEE international conference on
  computer vision. pp. 1742--1750 (2015)

\bibitem{pape20143}
Pape, J.M., Klukas, C.: 3-d histogram-based segmentation and leaf detection for
  rosette plants. In: European Conference on Computer Vision. pp. 61--74.
  Springer (2014)

\bibitem{quan2016fusionnet}
Quan, T.M., Hildebrand, D.G., Jeong, W.K.: Fusionnet: A deep fully residual
  convolutional neural network for image segmentation in connectomics. arXiv
  preprint arXiv:1612.05360  (2016)

\bibitem{ren17recattend}
Ren, M., Zemel, R.S.: End-to-end instance segmentation with recurrent
  attention. In: CVPR (2017)

\bibitem{romera2016recurrent}
Romera-Paredes, B., Torr, P.H.S.: Recurrent instance segmentation. In: European
  conference on computer vision. pp. 312--329. Springer (2016)

\bibitem{ronneberger2015u}
Ronneberger, O., Fischer, P., Brox, T.: U-net: Convolutional networks for
  biomedical image segmentation. In: International Conference on Medical image
  computing and computer-assisted intervention. pp. 234--241. Springer (2015)

\bibitem{scharr2016leaf}
Scharr, H., Minervini, M., French, A.P., Klukas, C., Kramer, D.M., Liu, X.,
  Luengo, I., Pape, J.M., Polder, G., Vukadinovic, D., et~al.: Leaf
  segmentation in plant phenotyping: a collation study. Machine vision and
  applications  \textbf{27}(4),  585--606 (2016)

\bibitem{silver2016mastering}
Silver, D., Huang, A., Maddison, C.J., Guez, A., Sifre, L., Van Den~Driessche,
  G., Schrittwieser, J., Antonoglou, I., Panneershelvam, V., Lanctot, M.,
  et~al.: Mastering the game of go with deep neural networks and tree search.
  nature  \textbf{529}(7587), ~484 (2016)

\bibitem{song2018seednet}
Song, G., Myeong, H., Mu~Lee, K.: Seednet: Automatic seed generation with deep
  reinforcement learning for robust interactive segmentation. In: Proceedings
  of the IEEE Conference on Computer Vision and Pattern Recognition. pp.
  1760--1768 (2018)

\bibitem{uhrig2016pixel}
Uhrig, J., Cordts, M., Franke, U., Brox, T.: Pixel-level encoding and depth
  layering for instance-level semantic labeling. In: German Conference on
  Pattern Recognition. pp. 14--25. Springer (2016)

\bibitem{zhang2016instance}
Zhang, Z., Fidler, S., Urtasun, R.: Instance-level segmentation for autonomous
  driving with deep densely connected {MRFs}. In: Proceedings of the IEEE
  Conference on Computer Vision and Pattern Recognition. pp. 669--677 (2016)

\bibitem{zhang2015monocular}
Zhang, Z., Schwing, A.G., Fidler, S., Urtasun, R.: Monocular object instance
  segmentation and depth ordering with {CNNs}. In: Proceedings of the IEEE
  International Conference on Computer Vision. pp. 2614--2622 (2015)

\end{thebibliography}

\end{document}